\documentclass[10pt,twocolumn,letterpaper]{article}

\usepackage{iccv}
\usepackage{times}
\usepackage{epsfig}
\usepackage{graphicx}
\usepackage{amsmath}
\usepackage{amssymb}
\usepackage{booktabs}
\usepackage{multirow}
\usepackage{arydshln}
\usepackage{flushend}
\usepackage{environ}
\usepackage[pagebackref=true,breaklinks=true,letterpaper=true,colorlinks,bookmarks=false]{hyperref}

\iccvfinalcopy % *** Uncomment this line for the final submission

\def\iccvPaperID{6324} % *** Enter the ICCV Paper ID here

\ificcvfinal\fi

\newcommand{\bboxrep}{\mathbf{B}}

\newcommand{\objectlabelrep}{\mathbf{O}}

\newcommand{\objectlabelrepeq}{$\objectlabelrepeq$}
\newcommand{\relationlabelrepeq}{$\relationlabelrepeq$}

\newcommand{\acksection}{\section*{Acknowledgments and Disclosure of Funding}}
\NewEnviron{ack}{%
  \acksection
  \BODY
}

\newcommand\blfootnote[1]{%
  \begingroup
  \renewcommand\thefootnote{}\footnote{#1}%
  \addtocounter{footnote}{-1}%
  \endgroup
}
\begin{document}

%%%%%%%%% TITLE
% \title{Semi-supervised Segmentation Grounded Scene Graph Generation}
\title{Segmentation-grounded Scene Graph Generation}

\author{
Siddhesh Khandelwal$^{*,1,2}$ \hspace{0.25in} Mohammed Suhail$^{*,1,2}$ \hspace{0.25in} Leonid Sigal$^{1,2,3}$\\
%  University of British Columbia\\
% \texttt{\{skhandel, rgoyal14, lsigal\}@cs.ubc.ca}\\
$^1$Department of Computer Science, University of British Columbia\\
% , Vancouver, BC, Canada \\
$^2$Vector Institute for AI \hspace{1in}   $^3$CIFAR AI Chair \\
\texttt{skhandel@cs.ubc.ca} \hspace{0.25in}
\texttt{suhail33@cs.ubc.ca} \hspace{0.25in}
\texttt{lsigal@cs.ubc.ca} \\
}

\maketitle
% Remove page # from the first page of camera-ready.
\ificcvfinal\thispagestyle{empty}\fi

%%%%%%%%% ABSTRACT
\begin{abstract}
% Scene graph generation has emerged as an important problem in computer vision. While scene graphs provide a grounded representation of objects, their locations and relations in an image, they do so only at the granularity of proposal bounding boxes. 
% In this work, we propose the first, to our knowledge, framework for pixel-level segmentation-grounded scene graph generation. 
% Our framework is agnostic to the underlying scene graph generation method and address the lack of segmentation annotations in target scene graph datasets (e.g., Visual Genome \cite{krishna2017visual}) through transfer and multi-task learning from, and with, an auxiliary dataset (e.g., MS COCO \cite{lin2014microsoft}). 
% Specifically, for a target object being detected, the segmentation mask is expressed as a lingual-similarity weighted linear combination over categories for which annotations are present in an auxiliary dataset. 
% To estimate these masks, object predictions are endowed with segmentation heads. 
% The relations prediction leverages object masks obtained using this segmentation head, along with a novel Gaussian attention mechanism which grounds the relations at a pixel-level within the image.
% The entire framework is end-to-end trainable and is learned in a multi-task manner with both target and auxiliary datasets. 

Scene graph generation has emerged as an important problem in computer vision. While scene graphs provide a grounded representation of objects, their locations and relations in an image, they do so only at the granularity of proposal bounding boxes. 
In this work, we propose the first, to our knowledge, framework for pixel-level segmentation-grounded scene graph generation. 
Our framework is agnostic to the underlying scene graph generation method and address the lack of segmentation annotations in target scene graph datasets (e.g., Visual Genome \cite{krishna2017visual}) through transfer and multi-task learning from, and with, an auxiliary dataset (e.g., MS COCO \cite{lin2014microsoft}). 
Specifically, each target object being detected is endowed with a segmentation mask, which is expressed as a lingual-similarity weighted linear combination over categories that have annotations present in an auxiliary dataset. These inferred masks, along with a novel Gaussian attention mechanism which grounds the relations at a pixel-level within the image, allow for improved relation prediction. The entire framework is end-to-end trainable and is learned in a multi-task manner with both target and auxiliary datasets. \blfootnote{$^*$Denotes equal contribution}

\end{abstract}
\begin{figure}
    \centering
    \includegraphics[width=0.37\textwidth]{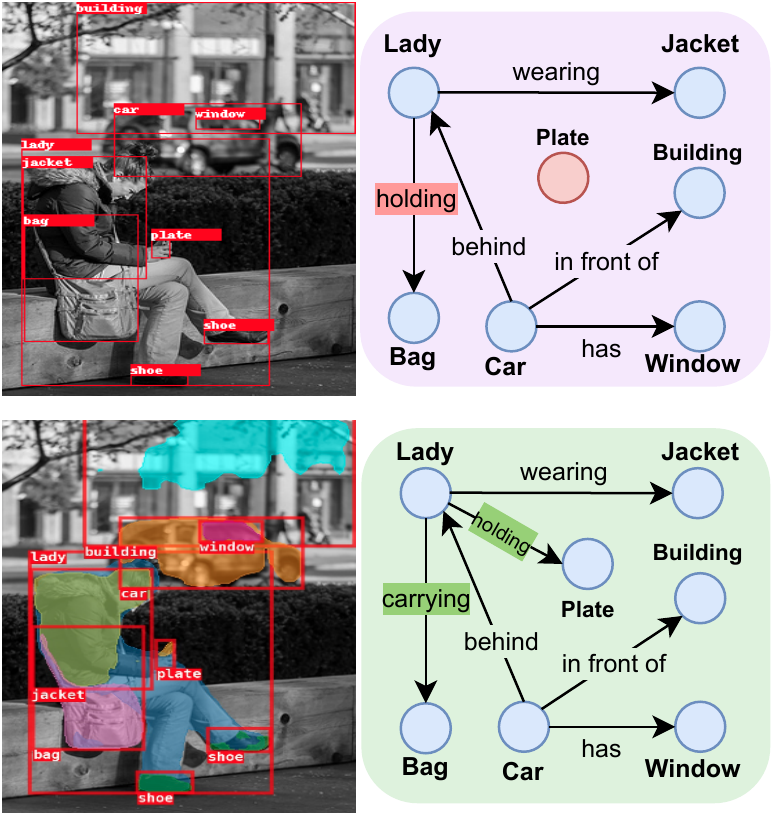}
    \caption{{\bf Segmentation-grounded scene graph generation.} The image on top is the output of an existing scene graph generation method \cite{tang2019learning}. The bottom image is the output of augmenting our approach to \cite{tang2019learning}. The effective grounding of objects to pixel-level regions within the image leads to better relation predictions.} 
    \label{fig:teaser}
    \vspace{-1.3em}
\end{figure}
%%%%%%%%% BODY TEXT
\section{Introduction}
Scene graph generation, or parsing, has emerged as a dominant problem in computer vision literature over the last couple of years \cite{newell2017pixels, tang2020unbiased, xu2017scene, yang2018graph, ZareianKC20}. The task involves producing a graph-based {\em grounded} representation of an image, which characterizes objects and their relationships. A scene graph representation, first introduced in \cite{xu2017scene}, encodes a scene in terms of a graph where nodes correspond to objects (encoding object instances with corresponding class labels and spatial locations) and directed edges corresponding to the predicate relationships. The ultimate goal of scene graph generation is to produce such representation from the raw image \cite{krishna2017visual} or video \cite{ji2020action}. Scene graph representations have proved to be important for a variety of higher level tasks (e.g., VQA\cite{hudson2018gqa,tang2019learning}, image captioning \cite{gu2019unpaired,yang2019auto} and others). Most approaches to date have focused on appropriate modeling of context \cite{yang2018graph,zellers2018neural}, data imbalance in labels \cite{Lin_2020_CVPR,tang2020unbiased} and, most recently, structural dependencies among the output variables \cite{suhail2021energybased}.

One of the dominant limitations of all existing scene graph generation techniques, mentioned above, is the fact that both the nodes (objects) and edges (relations) are grounded to (rectangular) bounding boxes produced by the object proposal mechanism directly (\eg, pre-trained as part of R-CNN) or by taking a union of bounding boxes of objects involved in a relation. A more granular and accurate pixel-level grounding would naturally be more valuable. This has been shown to be the case in other visual and visual-lingual tasks (\eg, referring expression comprehension  \cite{hu2016eccv,liu2017iccv}, video  segmentation with referring expression \cite{KhorevaACCV} and instance segmentation with Mask-RCNN \cite{he2017iccv}). In addition, grounding to segmentations could improve the overall performance of the scene graph generation by focusing node and edge features on irregular regions corresponding to objects or {\em interface} between objects, constituting an interaction. The goal of our approach is to do just that.

However, pixel-level ground comes with a number of unique challenges. The foremost of which is that traditional scene graph datasets, such as Visual Genome \cite{krishna2017visual}, do not come with instance-level segmentation annotations. This makes it impossible to employ a traditional fully supervised approach. 
% , \eg, in a way Mask RCNN was able to extend Faster RCNN with a segmentation mask and corresponding cross-entropy pixel-level loss. 
Further, even if we were to collect segmentation annotations, doing so for a large set of object types typically involved in Scene Graph predictions would be prohibitively expensive\footnote{As per \cite{bearman2016s}, labelling one image in VOC \cite{Everingham15} takes 239.7 seconds.}. To address this, we propose a transfer and multi-task learning formulation that uses an external dataset (\eg, MS COCO \cite{lin2014microsoft}) to provide segmentation annotations for some categories; while leveraging standard scene graph dataset (\eg, Visual Genome \cite{krishna2017visual}) to provide graph and bounding box annotations for the target % scene graph 
task.

% In doing so, we illustrate two core benefits. First, a zero-shot transfer mechanism allows us to produce segmentations for objects that have no segmentation annotations (in either dataset). Second, contextual feature refinement which is inherent to the scene graph predictions, improves on the original MS COCO segmentation performance. In other words, while providing qualitatively refined characterization of the task by showing ability to ground scene graphs at a pixel-level, as a byproduct, we show improvements on both scene graph prediction and instance-level segmentation tasks.

On a technical level, for a target object that lacks segmentation annotations, its mask is expressed as a weighted linear combination over categories for which annotations are present in an external dataset. This transfer is realised by leveraging the linguistic similarities between the target object label and these supervised categories, thus enabling grounding objects to segmentation masks without introducing any annotation cost. For a pair of objects that share a relation, our approach additionally employs a novel Gaussian attention mechanism to assign this relation to a pixel-level region within the image. Through joint optimization over the tasks of scene graph and segmentation generation, our approach achieves simultaneous improvements over both tasks.
Our proposed method is end-to-end trainable, and can be easily integrated into any existing scene graph generation method (\eg, \cite{tang2019learning,zellers2018neural}).

% On a technical level, our proposed method provides an end-to-end solution to grounding scene graphs to segmentations, and can be easily integrated to existing scenegraph methods like \cite{tang2019learning,zellers2018neural}. For a pair of objects, these methods rely on using features computed over the union of the corresponding bounding boxes for relation prediction. Our proposed approach augments this feature computation procedure with a novel Gaussian attention mechanism to incorporate segmentation annotations to ground relations to specific pixel-level regions in the image. These segmentation annotations, which are computed for each object via a \emph{zero-shot} transfer mechanism, analogously enable grounding objects to pixel-level regions without introducing any additional labelling cost.

\vspace{0.4em}
\noindent
{\bf Contributions:}
% Our foremost contribution is that we propose the first, to our knowledge, framework for pixel-level segmentation-grounded scene graph generation/prediction. 
% This framework is agnostic to the underlying scene graph generation method and addresses the lack of segmentation annotations in target scene graph datasets (\eg, Visual Genome \cite{krishna2017visual}) through transfer and multi-task learning from, and with, an auxiliary dataset (\eg, MS COCO \cite{lin2014microsoft}). 
% Specifically, each detected target object is endowed with a segmentation mask, which is expressed as a lingual-similarity weighted linear combination over categories that have annotations present in an auxiliary dataset. These inferred masks, along with a novel Gaussian attention mechanism, which grounds the relations at a pixel-level within the image, allow for improved relation prediction. 
Our foremost contribution is that we propose the first, to our knowledge, framework for pixel-level segmentation-grounded scene graph generation/prediction, which can be integrated with any existing scene graph generation method. For objects, these groundings are realised via segmentation masks, which are computed through a lingual-similarity based zero-shot transfer mechanism over categories in an auxiliary dataset. To effectively ground relations at a pixel level, we additionally propose a novel Gaussian attention mechanism over segmentation masks. Finally, we demonstrate the flexibility and efficacy of our approach by augmenting it to existing scene graph architectures, and evaluating performance on the Visual Genome \cite{krishna2017visual} benchmark dataset, where we consistently outperform baselines by up to $12\%$ on relation prediction.
% Through jointly optimizing over the tasks of scene graph and segmentation generation, we refine the inferred masks while simultaneously further improving on relation prediction. 
\vspace{-0.1em}

\begin{figure*}[t]
    \centering
    \includegraphics[width=\linewidth]{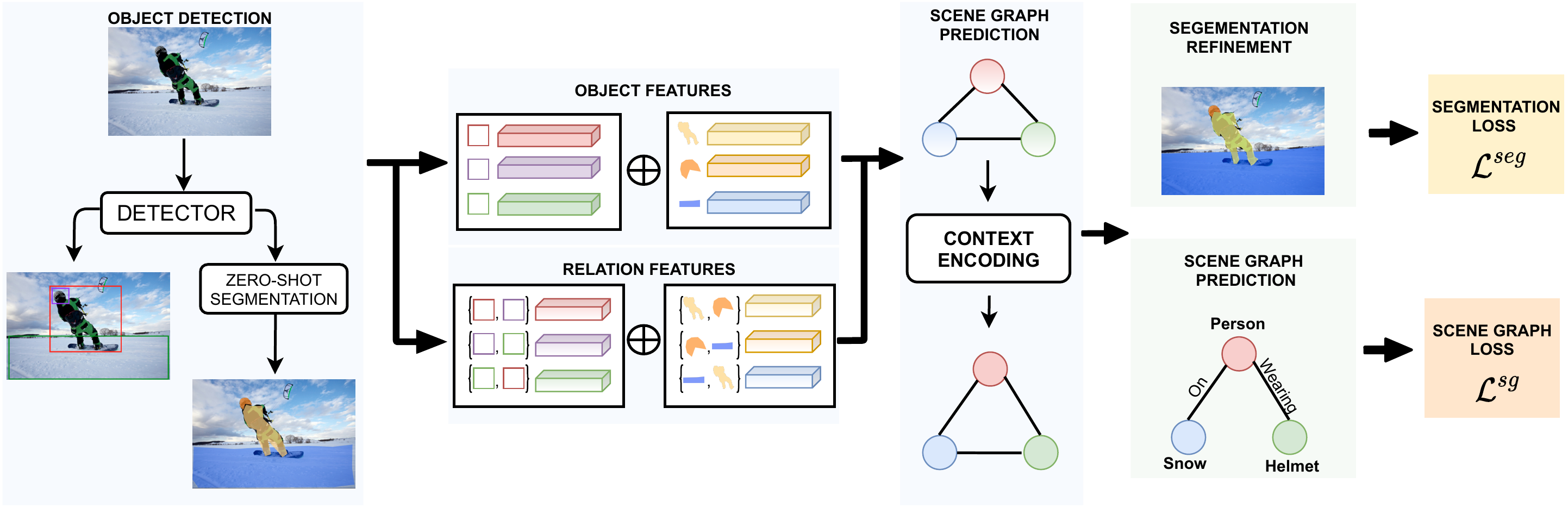}
    \caption{\textbf{Model Architecture.} For an image, the object detector provides a set of bounding boxes, and for each box, additionally generates instance-level segmentations via a \emph{zero-shot} transfer mechanism. These inferred segmentation masks are incorporated into the nodes and edges of the underlying graph, before passing it into an existing scene graph prediction architecture like \cite{zellers2018neural,tang2019learning}. The inferred segmentation masks are additionally refined by leveraging the global context captured by the context aggregation step of the scene graph prediction method. The proposed method is end-to-end trainable, and can be augmented to any existing scene graph method.
    }
    \label{fig:model_overview}
    \vspace{-1.5em}
\end{figure*}
\vspace{-0.4em}
\section{Approach}
\label{sec:approach}
\vspace{-0.4em}
We propose a novel multi-task learning framework that leverages instance-level segmentation annotations, obtained via a \emph{zero-shot} transfer mechanism, to effectively generate %granular 
pixel-level groundings for the objects within a scene graph. Our approach, highlighted in Figure \ref{fig:model_overview}, builds on existing scene graph generation methods, but is \emph{agnostic} to the underlying architecture and can be easily integrated with existing state-of-the-art approaches. 

\vspace{-0.4em}
\subsection{Notation}
\vspace{-0.4em}
% Consistent with existing scene graph generation literature \cite{tang2019learning,zellers2018neural},  we 
Let $\mathcal{D}^{g} = \{ (\mathbf{x}^{g}_i, \mathbf{G}^{g}_i) \}$ denote the dataset containing graph-level annotations $\mathbf{G}^{g}_i$ for each image $\mathbf{x}^{g}_i$. We represent the scene graph annotation $\mathbf{G}^{g}_i$ as a tuple of object and relations, $\mathbf{G}^{g}_i = (\mathbf{O}^{g}_i, \mathbf{R}^{g}_i)$, where $\mathbf{O}^{g}_i \in \mathbb{R}^{n_i \times d_g}$ represents object labels and $\mathbf{R}^{g}_i \in \mathbb{R}^{n_i \times n_i \times d_g'}$ represents relationship labels; $n_i$ is the number of objects in an image $\mathbf{x}^{g}_i$; $d_g$ and $d_g'$ are the total number of possible object and relation labels, respectively, in the dataset.

In addition, we assume availability of the dataset $\mathcal{D}^{m} = \{(\mathbf{x}^{m}_i, \mathbf{M}^{m}_i)\}$, where each image $\mathbf{x}^{m}_i$ has corresponding instance-level segmentation annotations $\mathbf{M}^{m}_i$. Finally, $d_m$ are the total number of possible object labels in $\mathcal{D}^{m}$.

As is the case with existing scene graph datasets like Visual Genome \cite{krishna2017visual}, $\mathcal{D}^{g}$ does not contain any instance-level segmentation masks. Also, $\mathcal{D}^{m}$ can be any dataset (like MS COCO \cite{lin2014microsoft}).
Note, that in general, the images in the two datasets, $\mathcal{D}^{g}$ and $\mathcal{D}^{m}$, are disjoint and the object classes in the two datasets may have minimal overlap (\eg, MS COCO provides segmentations for 80, while Visual Genome provides object bounding boxes for 150 object categories\footnote{Visual Genome has a ${\sim}47\%$ image overlap with MS-COCO. However, they have differing object categories and annotations. We make no use of this implicit image overlap in our formulation.}). 

% , and can be disjoint from $\mathcal{D}^{g}$. 
For brevity, we drop subscript $i$ for the rest of the paper. 

\vspace{-0.3em}
\subsection{Scene Graph Generation}
\vspace{-0.3em}
\label{sec:sg-generation}
Given an image $\mathbf{x}^{g} \in \mathcal{D}^{g}$, a typical scene graph model defines the distribution over the scene graph $\mathbf{G}^{g}$ as follows,
\begin{align}
    \label{eq:sg-factor}
    \text{Pr}\left(\mathbf{G}^{g}| \mathbf{x}^{g} \right) = \text{Pr}\left(\bboxrep^{g}|\mathbf{x}^{g} \right)\cdot
    \text{Pr}\left(\mathbf{O}^{g} |\bboxrep^{g},\mathbf{x}^{g} \right)\cdot
    \text{Pr}\left(\mathbf{R}^{g}|\mathbf{O}^{g},\bboxrep^{g},\mathbf{x}^{g} \right)
\end{align}
The bounding box network $\text{Pr}\left(\bboxrep^{g}|\mathbf{x}^{g} \right)$ extracts a set of boxes $\bboxrep^{g} = \{\mathbf{b}^{g}_{1}, \dots, \mathbf{b}^{g}_{n} \}$ corresponding to regions of interest. This can be achieved using standard object detectors such as Faster R-CNN \cite{ren2015faster} or Detectron \cite{wu2019detectron2}. Specifically, these detectors are pretrained on $\mathcal{D}^{g}$ with the objective to generate accurate bounding boxes $\mathbf{B}^b$ and object probabilities $\mathbf{L}^g = \{\mathbf{l}^g_1, \dots, \mathbf{l}^g_n\}$ for an input image $\mathbf{x}^{g}$. Note that this only requires access to the object (node) annotations in $\mathbf{G}^{g}$.

The object network $\text{Pr}\left(\mathbf{O}^{g} |\bboxrep^{g},\mathbf{x}^{g} \right)$, for each bounding box $\mathbf{b}^{g}_{j} \in \bboxrep^{g}$, utilizes feature representation $\mathbf{z}^{g}_{j}$, where $\mathbf{z}^{g}_{j}$ is computed as $\text{\tt RoIAlign}(\mathbf{x}^{g}, \mathbf{b}^{g}_{j})$, which extracts features from the area within the image corresponding to the bounding box $\mathbf{b}^{g}_{j}$. These features, alongside object label probabilities $\mathbf{l}^g_j$,
% \leon{we never talk about how we get these and probably should},
are fed into a context aggregation layer such as Bi-directional LSTM \cite{zellers2018neural}, Tree-LSTM \cite{tang2019learning}, or Graph Attention Network \cite{yang2018graph}, to obtain refined features $\mathbf{z}^{o,g}_j$. 
% Features representation $\mathbf{z}^{g}_{i,j}$ for a specific bounding box $\mathbf{b}^{g}_{i,j} \in \bboxrep^{g}_{i}$ is computed as $\mathbf{z}^{g}_{i,j} = \text{\tt RoIAlign}(\mathbf{x}^{g}_i, \mathbf{b}^{g}_{i,j})$. These representation along with an initial prediction of the object labels are are used as input to a context aggregation layer such as Bi-directional LSTM \cite{zellers2018neural}, Tree-LSTM \cite{tang2019learning}, Graph Attention Network \cite{yang2018graph} \etc.
These refined features are used to obtain the object labels $\objectlabelrep^g$ for the nodes within the graph $\mathbf{G}^{g}$.
% via a prediction network $f_{\mathbf{N}}(.)$. 

Similarly, for the relation network $\text{Pr}\left(\mathbf{R}^{g}|\mathbf{O}^{g},\bboxrep^{g},\mathbf{x}^{g} \right)$, features corresponding to union of object bounding boxes are refined using message passing layers and subsequently classified to produce predictions for relations.
% using a relation network $f_{\mathbf{E}}(.)$.

% These refined features used to obtain the object labels $\objectlabelrep$. 
% To obtain the relation labels ($\relationlabelrep$), the model first extracts features for regions corresponding to union of object bounding boxes for different object pairs. These feature are further refined using message passing layers and subsequently classified to obtain relation labels.
Existing models ground the objects in the scene graph to rectangular regions in the image. While grounding with bounding boxes provides an approximate estimate of the object locations, having a more granular pixel-level grounding achievable through segmentation masks is much more desirable. A major challenge is the lack of segmentation annotation in scene graph datasets like Visual Genome \cite{krishna2017visual}. Furthermore, manually labelling segmentation masks for such large datasets is both time consuming and expensive. As a solution, we derive segmentation masks via a \emph{zero-shot} transfer mechanism from a segmentation head trained on an external dataset $\mathcal{D}^{m}$(\eg MSCOCO \cite{lin2014microsoft}). This inferred segmentation mask is then used as additional input to the object and relation networks to generate better scene graphs. 
% Specifically, our approach factorizes the distribution over scene graphs $\mathbf{G}^{g}$ as follows, 
Our approach factorizes the distribution over $\mathbf{G}^{g}$ as,
\begin{align}
    \label{eq:sg-seg-factor}
    \begin{split}
    \text{Pr}\left(\mathbf{G}^{g}| \mathbf{x}^{g} \right) = \text{Pr}\left(\bboxrep^{g}|\mathbf{x}^{g} \right)\cdot &\text{Pr}\left(\mathbf{M}^{g}|\mathbf{x}^{g} \right)\cdot
    \text{Pr}\left(\mathbf{O}^{g} |\bboxrep^{g},\mathbf{M}^{g},\mathbf{x}^{g} \right)\\
    \cdot&\text{Pr}\left(\mathbf{R}^{g}|\mathbf{O}^{g},\bboxrep^{g},\mathbf{M}^{g},\mathbf{x}^{g} \right)  
    \end{split}
\end{align}
where $\mathbf{M}^{g} = \{\mathbf{m}^g_{1}, \dots, \mathbf{m}^g_{n}\}$ are the inferred segmentation masks corresponding to the bounding boxes $\mathbf{B}^g$. Such a factorization enables grounding scene graphs to segmentation masks and affords easy integration to existing architectures.
%In the next sections we describe the individual components of this factorization in detail.

\vspace{-0.3em}
\subsection{Segmentation Mask Transfer}
\vspace{-0.3em}
\label{sec:sg-transfer}
% As described in Equation \ref{eq:sg-seg-factor}, for each object bounding box, our proposed approach relies on the availability of segmentation masks for improved scene graph generation. However, existing popular scene graph datasets like Visual Genome \cite{krishna2017visual} lack segmentation annotations. Furthermore, manually labelling segmentation masks for such large datasets is both time consuming and expensive.
% % \footnote{As per \cite{bearman2016s}, labelling one image in VOC \cite{Everingham15} takes 239.7 seconds.}.
% To alleviate this issue, our approach utilizes segmentation annotations from existing publicly available datasets such as MSCOCO \cite{lin2014microsoft} to obtain segmentation net  $\text{Pr}\left(\mathbf{M}^{g}| \mathbf{x}^{g}\right)$. 

For each image $\mathbf{x}^{g} \in \mathcal{D}^{g}$, we derive segmentation masks $\mathbf{M}^{g}$ using annotations learned over classes in an external dataset $\mathcal{D}^m$. To facilitate this, like described in Section \ref{sec:sg-generation}, we pretrain a standard object detector (like Faster R-CNN \cite{ren2015faster}) on the scene graph dataset $\mathcal{D}^{g}$. However, instead of training the detector just on images in $\mathcal{D}^{g}$, we additionally jointly learn a segmentation head $f_{\mathbf{M}}$ on images in $\mathcal{D}^m$. 
Note that when training the object detector jointly on images in $\mathcal{D}^{g}$ and $\mathcal{D}^{m}$, the same backbone and proposal generators are used, thus reducing the memory overhead. 
% Note that both the object detector trained on $\mathcal{D}^{g}$ and the segmentation head trained on $\mathcal{D}^{m}$ share the same backbone and proposal generator, thus significantly reducing the memory. %  overhead.

For an image $\mathbf{x}^{g} \in \mathcal{D}^{g}$,  let $\mathbf{z}^{g}_{j}$ be the feature representation for a bounding box $\mathbf{b}^{g}_{j} \in \bboxrep^{g}$. Let,
% $\widetilde{\mathbf{m}}^{g}_{j}$ be defined as,
\begin{align}
    \widetilde{\mathbf{m}}^{g}_{j}  = f_{\mathbf{M}}(\mathbf{z}^{g}_{j})
\end{align}
where $\widetilde{\mathbf{m}}^{g}_{j} \in \mathbb{R}^{d_{m} \times m \times m}$, $d_{m}$ represents the number of classes in $\mathcal{D}^{m}$, and $m$ is the spatial resolution of the mask. Per class segmentation masks ${\mathbf{m}}^{g}_{j} \in \mathbb{R}^{d_{g} \times m \times m}$ are then derived from $\widetilde{\mathbf{m}}^{g}_{j}$ using a \emph{zero-shot} transfer mechanism\footnote{Note that $d_g >> d_m$ in our case.}. Let $\mathbf{S} \in \mathbb{R}^{d_g \times d_m}$ be a matrix that captures linguistic similarities between classes in $\mathcal{D}^{g}$ and $\mathcal{D}^{m}$. For a pair of classes $c_g \in [1, d_g], c_m \in [1, d_m]$, the element $\mathbf{S}_{c_g,c_m}$ is defined as,
\begin{align}
    \mathbf{S}_{c_g,c_m} = \mathbf{g}_{c_g}^\top\mathbf{g}_{c_m}
\end{align}
where $\mathbf{g}_{c_g}$ and $\mathbf{g}_{c_m}$ are $300$-dimensional GloVe \cite{pennington2014glove} vector embeddings for classes $c_g$ and $c_m$ respectively\footnote{For class names that contain multiple words, individual GloVe word embeddings are averaged.}. ${\mathbf{m}}^{g}_{j}$ is then obtained as a linear combination over $\widetilde{\mathbf{m}}^{g}_{j}$ as follows,
\begin{align}
    {\mathbf{m}}^{g}_{j} = \mathbf{S}^\top \widetilde{\mathbf{m}}^{g}_{j}
\end{align}
Note that such a transfer doesn't require \emph{any} additional labelling cost as we rely on a publicly available dataset $\mathcal{D}^m$.
\vspace{-0.3em}
\subsection{Grounding Nodes to Segmentation Masks}
\vspace{-0.3em}
\label{sec:sg-object}
As mentioned in Equation \ref{eq:sg-seg-factor}, we incorporate the inferred segmentation masks in the object network $\text{Pr}\left(\mathbf{O}^{g} |\bboxrep^{g},\mathbf{M}^{g},\mathbf{x}^{g}\right)$ to ground objects in $\mathcal{D}^g$ to pixel-level regions within the image.

Specifically, for a particular image $\mathbf{x}^g$, the model $\text{Pr}\left(\mathbf{B}^g|\mathbf{x}^g \right)$ outputs a set of bounding boxes $\mathbf{B}^g$. For each bounding box $\mathbf{b}^g_{j} \in \mathbf{B}^g$, it additionally also computes a feature representation $\mathbf{z}^g_{j}$ and object label probabilities $\mathbf{l}^g_{j} \in \mathbb{R}^{d_g + 1}$ (includes background as a possible label). Following the procedure described in Section \ref{sec:sg-transfer}, per-class segmentation masks $\mathbf{m}^g_{j}$ are inferred for each bounding box $\mathbf{b}^g_{j}$. We define a segmentation aware representation $\hat{\mathbf{z}}^g_{j}$ as,
\begin{align}
    \hat{\mathbf{z}}^g_{j} = f_{\mathbf{N}}\left(~[{\mathbf{z}}^g_{j}, \mathbf{m}^g_{j}]~\right)
\end{align}
where $f_{\mathbf{N}}$ is a learned network and $[.,.]$ represents concatenation. Contrary to existing methods like \cite{tang2019learning,zellers2018neural} that use the segmentation agnostic representation $\mathbf{z}^g_{j}$, we feed $\hat{\mathbf{z}}^g_{j}$ and $\mathbf{l}^g_{j}$ as inputs to the object network ${\text{Pr}\left(\mathbf{O}^{g} |\bboxrep^{g},\mathbf{M}^{g},\mathbf{x}^{g}\right)}$. 
\vspace{-0.3em}
\subsection{Grounding Edges to Segmentation Masks}
\vspace{-0.3em}
\label{sec:sg-relation}
To facilitate better relation prediction, we leverage the inferred segmentation masks in the relation network $\text{Pr}\left(\mathbf{R}^{g}|\mathbf{O}^{g},\bboxrep^{g},\mathbf{M}^{g},\mathbf{x}^{g}\right)$. Specifically, for a pair of objects, we utilize a novel Gaussian attention mechanism to identify relation identifying pixel-level regions within an image.

Given a pair of bounding boxes $(\mathbf{b}^g_{j}, \mathbf{b}^g_{j'}) \in \mathbf{B}^g$ that contain a possible edge and their corresponding object label probabilities $(\mathbf{l}^g_j, \mathbf{l}^g_{j'})$, their respective segmentation masks $(\mathbf{m}^g_{j}, \mathbf{m}^g_{j'})$ are computed via the procedure described in Section \ref{sec:sg-transfer}. We define $\mathbf{z}^g_{j,j'}$ as the segmentation agnostic feature representation representing the union of boxes $(\mathbf{b}^g_{j}, \mathbf{b}^g_{j'})$, which is computed as $\texttt{RoIAlign}(\mathbf{x}^g, \mathbf{b}^g_{j} \cup \mathbf{b}^g_{j'} )$\footnote{$\mathbf{b}^g_{j} \cup \mathbf{b}^g_{j'}$ computes the convex hull of the union of the two boxes.}.

Contrary to existing works that rely on this coarse rectangular union box, our approach additionally incorporates a union of the segmentation masks $(\mathbf{m}^g_{j}, \mathbf{m}^g_{j'})$ to provide more granular information. To this end, we define the attended union segmentation mask $\mathbf{m}^g_{j,j'}$ as,
\begin{align}
    \mathbf{m}^g_{j,j'} = (\mathbf{K}_j \circledast \mathbf{m}^g_{j}) \odot (\mathbf{K}_{j'} \circledast \mathbf{m}^g_{j'})
\end{align}
where $\circledast$ is the convolution operation, and $\odot$ computes an element-wise product. $\mathbf{K}_j, \mathbf{K}_{j'}$ are $\delta\times\delta$ sized Gaussian smoothing spatial convolutional filters
parameterized by variances $\sigma_x^2$, $\sigma_y^2$ and
correlation $\rho_{x,y}$.
%. Specifically, a particular element $k_{a,b} \in \mathbf{K}_j; a,b\in[1,\delta]$ is defined as,
%\begin{align}
%    \begin{split}
%       k_{a,b} &= \frac{1}{2\pi\sigma_x\sigma_y\sqrt{1 - \rho_{x,y}^2}}\exp\left[-\frac{t_{a,b}}{2(1 - \rho_{x,y}^2)}\right] \\
%       t_{a,b} &= \frac{(a - \frac{\delta}{2})^2}{\sigma_x^2} - \frac{2\rho_{x,y}(a - \frac{\delta}{2})(b - \frac{\delta}{2})}{\sigma_x\sigma_y} +\frac{(b - \frac{\delta}{2})^2}{\sigma_y^2}
%    \end{split}
%\end{align}
% where $\sigma_x^2, \sigma_y^2$ are dimensional 
%variances, %values, 
% and $\rho_{x,y}$ represents correlation. 
These parameters are obtained by learning a transformation over the object label probabilities $\mathbf{l}^g_j$. Specifically,
\begin{align}
    \sigma_x^2, \sigma_y^2, \rho_{x,y} = f_{\mathcal{N}}\left(\mathbf{l}^g_j\right)
\end{align}
where $f_{\mathcal{N}}$ is a learned network. $\mathbf{K}_{j'}$ is computed analogously using $\mathbf{l}^g_{j'}$. The attended union segmentation mask $\mathbf{m}^g_{j,j'}$ affords the computation of a segmentation aware representation $\hat{\mathbf{z}}^g_{j,j'}$ as follows,
\begin{align}
    \hat{\mathbf{z}}^g_{j,j'} = f_{\mathbf{E}}\left([\mathbf{z}^g_{j,j'},  \mathbf{m}^g_{j,j'}]\right)
\end{align}
where $f_{\mathbf{E}}$ is a learned network.
% , and $[.,.]$ represents concatenation. 
$\hat{\mathbf{z}}^g_{j,j'}$ is then used as an input to the relation network ${\text{Pr}\left(\mathbf{R}^{g}|\mathbf{O}^{g},\bboxrep^{g},\mathbf{M}^{g},\mathbf{x}^{g}\right)}$.

\subsection{Refining Segmentation Masks}
\vspace{-0.3em}
\label{sec:sg-seg-refine}
% \textcolor{red}{TODO: Not sure how to position additional segmentation head.}
As described previously, our proposed approach incorporates segmentation masks to improve relation prediction. However, we posit that the tasks of segmentation and relation prediction are indelibly connected, wherein an improvement in one leads to an improvement in the other.

To this end, for each object $\mathbf{b}^g_j \in \mathbf{B}^g $, in addition to predicting the object labels $\mathbf{O}^g$, we learn a segmentation \emph{refinement} head $f_{\mathbf{M}'}$ to refine the inferred segmentation masks $\mathbf{m}^g_j$. However, as the scene graph dataset $\mathcal{D}^g$ does not contain any instance-level segmentation annotations, training $f_{\mathbf{M}'}$ in a traditionally supervised manner is challenging. 

To alleviate this issue, we again leverage the auxiliary dataset $\mathcal{D}^m$, which contains instance-level segmentation annotations. For an image $\mathbf{x}^m \in \mathcal{D}^m$, bounding boxes $\mathbf{B}^m$ are computed using the object detector. Note that this does not require any additional training as the object detector is jointly trained using both $\mathcal{D}^g$ and $\mathcal{D}^m$ as described in Section \ref{sec:sg-transfer}.
% \leon{I am not quite sure we say we train a detector on both datasets in Section \ref{sec:sg-transfer}. We say we train segmentation head on COCO. Should we clarify?}. 
For a bounding box $\mathbf{b}^m_j \in \mathbf{B}^m$, the corresponding per class masks are computed as,
\begin{align}
    \mathbf{m}^m_j = f_{\mathbf{M}}\left(\mathbf{z}^m_j \right)
\end{align}
where $\mathbf{z}^m_j$ is the feature representation for $\mathbf{b}^m_j$, and $f_{\mathbf{M}}$ is the segmentation head defined in Section \ref{sec:sg-transfer}. The refined mask $\hat{\mathbf{m}}^m_j$ is then computed as,
\begin{align}
    \hat{\mathbf{m}}^m_j = \mathbf{m}^m_j + f_{\mathbf{M}'}\left(\mathbf{z}^{o,m}_j \right)
\end{align}
where $\mathbf{z}^{o,m}_j$ is the representation computed by the context aggregation layer within the object network $\text{Pr}\left(\mathbf{O}^{m} |\bboxrep^{m},\mathbf{M}^{m},\mathbf{x}^{m} \right)$. Note that this network is identical to the one defined in Equation \ref{eq:sg-seg-factor}. The segmentation \emph{refinement} head $f_{\mathbf{M}'}$ is a zero-initialized network that learns a residual update over the mask $\mathbf{m}^m_j$. As ground-truth segmentation annotations are available for all objects $\mathbf{B}^m$, $f_{\mathbf{M}'}$ is trained using a pixel-level cross entropy loss. 

$f_{\mathbf{M}'}$ is trained alongside the scene graph generation model, and the refined masks are used during inference to improve relation prediction performance. Specifically, for a particular image $\mathbf{x}^g \in \mathcal{D}^g$, we follow the model described in Equation \ref{eq:sg-seg-factor} to generate predictions. However, instead of directly using the inferred masks obtained using the zero-shot formulation in Section \ref{sec:sg-transfer}, we additionally refine it using $f_{\mathbf{M}'}$. For a particular mask $\mathbf{m}^g_j$ corresponding to a bounding box $\mathbf{b}^g_j$, we compute $\hat{\mathbf{m}}^g_j$ as,
\begin{align}
    \hat{\mathbf{m}}^g_j = \mathbf{m}^g_j + f_{\mathbf{M}'}\left(\mathbf{z}^{o,g}_j \right)
\end{align}
where $\mathbf{z}^{o,g}_j$ is the representation computed by the context aggregation layer. The refined mask is used in the object and relation networks as described in Sections \ref{sec:sg-object} and \ref{sec:sg-relation}.

\subsection{Training}
\vspace{-0.3em}
Our proposed approach is trained in two stages. The first stage involves pre-training the object detector to enable bounding box proposal generation for a given image. Given datasets $\mathcal{D}^g$ and $\mathcal{D}^m$, the object detector is jointly trained to minimize the following objective,
\begin{align}
    \label{eq:pretrain}
    \mathcal{L}^{obj} = \mathcal{L}^{rcnn} + \mathcal{L}^{seg}
\end{align}
where $\mathcal{L}^{rcnn}$ is the Faster R-CNN \cite{ren2015faster} objective, and $\mathcal{L}^{seg}$ is the pixel-level binary cross entropy loss \cite{he2017iccv} applied over segmentation masks. Note that images in $\mathcal{D}^g$ do not contribute to $\mathcal{L}^{seg}$ due to lack of segmentation annotations. 

The second stage of training involves training the scene graph generation network to accurately identify relations between pairs of objects. Given datasets $\mathcal{D}^g$ and $\mathcal{D}^m$, the scene graph generation network is jointly trained to minimize the following objective,
\begin{align}
    \mathcal{L} = \mathcal{L}^{sg} + \mathcal{L}^{seg}
\end{align}
where $\mathcal{L}^{sg}$ depends on the architecture of the underlying scene graph method our approach is augmented to. For example, in the case of MOTIF \cite{zellers2018neural}, $\mathcal{L}^{sg}$ consists of two cross-entropy losses, one to refine the object categorization obtained from the pretrained detector, and the other to aide with accurate relation prediction. $\mathcal{L}^{seg}$ is identical to the segmentation loss described in Equation \ref{eq:pretrain}, and is used to learn the refinement network $f_{\mathbf{M}'}$ (Section \ref{sec:sg-seg-refine}). As images in $\mathcal{D}^m$ do not contain scene graph annotations, they only contribute to $\mathcal{L}^{seg}$. Similarly, images in $\mathcal{D}^m$ only affect $\mathcal{L}^{sg}$.

\renewcommand{\arraystretch}{1.2}
\begin{table*}[t]
\centering
\resizebox{\textwidth}{!}{
\begin{tabular}{@{}cccccccccccc@{}}
\toprule
\toprule
\multirow{2}{*}{Model}   &\multirow{2}{*}{Detector} & \multirow{2}{*}{Method}  & \multicolumn{3}{c}{Predicate Classification}     & \multicolumn{3}{c}{Scene Graph Classification}  & \multicolumn{3}{c}{Scene Graph Generation}    \\ \cmidrule(l){4-12} 
                         &    &                   & mR@20          & mR@50          & mR@100         & mR@20         & mR@50          & mR@100         & mR@20         & mR@50         & mR@100        \\ \midrule
IMP \cite{xu2017scene}                 & VGG-16 \cite{vgg-16}   & -                    & -              & 9.8            & 10.5           & -             & 5.8            & 6.0            & -             & 3.8           & 4.8           \\
% FREQ \cite{zellers2018neural}                     & -       & -                 & 8.3            & 13.0           & 16.0           & 5.1           & 7.2            & 8.5            & 4.5           & 6.1           & 7.1           \\
MOTIF \cite{zellers2018neural}                   & VGG-16 \cite{vgg-16}      & -                  & 10.8           & 14.0           & 15.3           & 6.3           & 7.7            & 8.2            & 4.2           & 5.7           & 6.6           \\
VCTree \cite{tang2019learning}                   & VGG-16 \cite{vgg-16}       & -                 & 14.0           & 17.9           & 19.4           & 8.2           & 10.1           & 10.8           & 5.2           & 6.9           & 8.0           \\ \midrule
\multirow{4}{*}{$\text{MOTIF}^\dagger$} & \multirow{2}{*}{VGG-16 \cite{vgg-16}}  & Baseline                & 13.7           & 17.5          & 18.9          & 7.5          & 9.2           & 9.8          & 5.2          & 6.8          & 7.9          \\
                                        &  & Seg-Grounded   & \textbf{14.6}  & \textbf{18.7}  & \textbf{20.3}  & \textbf{7.9} & \textbf{9.8}  & \textbf{10.5} & \textbf{5.6}  & \textbf{7.3}  & \textbf{8.1}  \\ \cdashline{3-12}
                                        & \multirow{2}{*}{\shortstack{ResNeXt-101- \\FPN \cite{massa2018mrcnn, xie2017aggregated}}}  & Baseline                & 14.1           & 18.0          & 19.4          & 8.0          & 9.9           & 10.6          & 5.8          & 7.7          & 9.0          \\
                                        &  & Seg-Grounded             & \textbf{14.5} & \textbf{18.5} & \textbf{20.2}  & \textbf{8.9} & \textbf{11.2} & \textbf{12.1} & \textbf{6.4} & \textbf{8.3} & \textbf{9.2} \\ \cmidrule(l){2-12} 
\multirow{4}{*}{$\text{VCTree}^\dagger$} 
                        & \multirow{2}{*}{VGG-16 \cite{vgg-16}}  & Baseline                & 14.4          & 18.4          & 19.8          & 8.1          & 9.9           & 10.7          & 4.4          & 5.7         &  6.4        \\
                       &   & Seg-Grounded            & \textbf{14.8}  & \textbf{18.9} & \textbf{20.5} & \textbf{8.7} & \textbf{10.8} & \textbf{11.6} & \textbf{5.3} & \textbf{7.0} & \textbf{7.8} \\\cdashline{3-12}

                        & \multirow{2}{*}{\shortstack{ResNeXt-101- \\FPN \cite{massa2018mrcnn, xie2017aggregated}}}  & Baseline                & 13.7          & 17.4          & 19.0          & 8.1          & 9.9           & 10.6          & 5.3          & 6.9          & 7.9          \\
                       &   & Seg-Grounded            & \textbf{15.0}  & \textbf{19.2} & \textbf{21.1} & \textbf{9.3} & \textbf{11.6} & \textbf{12.3} &  \textbf{6.3}             &  \textbf{8.1}             &     \textbf{9.0}          \\
\bottomrule
\bottomrule
\end{tabular}
}
\caption{\textbf{Scene Graph Prediction on Visual Genome. }Mean Recall (mR) is reported for three tasks, across two detector backbones. Our approach is augmented to and contrasted against MOTIF \cite{zellers2018neural} and VCTree \cite{tang2019learning}. $\dagger$ denotes our re-implementation of the methods.}
\label{table:results}
\vspace{-0.5em}
\end{table*}

\renewcommand{\arraystretch}{1.1}
\begin{table*}[t]
\setlength\tabcolsep{2.5pt}
\resizebox{\linewidth}{!}{
\begin{tabular}{@{}cc@{\hskip 0.2in}cccccc@{\hskip 0.2in}cccccc@{\hskip 0.2in}cccccc@{}}
\toprule
\toprule
 \multirow{2}{*}{Detector} & \multirow{2}{*}{Method} & \multicolumn{6}{{c@{\hskip 0.2in}}}{Predicate Classification}                & \multicolumn{6}{c@{\hskip 0.2in}}{Scene Graph Classification}              & \multicolumn{6}{c}{Scene Graph Generation}                  \\\cmidrule(l){3-20} 
                         &                         & AP & AP$_{50}$ & AP$_{75}$ & AP$_{S}$ & AP$_{M}$ & AP$_{L}$ & AP & AP$_{50}$ & AP$_{75}$ & AP$_{S}$ & AP$_{M}$ & AP$_{L}$ & AP & AP$_{50}$ & AP$_{75}$ & AP$_{S}$ & AP$_{M}$ & AP$_{L}$ \\\midrule
\multirow{3}{*}{VGG-16 \cite{vgg-16}}      & No Refine                     & 31.5   & 63.8          & 28.1           & 21.8          & 36.4          & 43.9          & 32.5    & 58.9           & 31.8           & 17.0          & 35.3         & 42.3          & 23.2   & 44.7          & 21.6          & 8.1         &    26.0      & 35.1         \\
                          & MOTIF$^\dagger + \text{Refine}$    & \textbf{42.4}   & \textbf{78.1}          & \textbf{40.9}          & \textbf{33.0}          & \textbf{46.6}         & \textbf{55.8}          &    \textbf{37.5} &    \textbf{63.5}       &    \textbf{38.8}       &    \textbf{21.0}      & \textbf{40.7}         & \textbf{48.4}          & 24.7   &  45.8         &  23.9         &  8.6        & 27.9          & 38.1          \\
                          & VCTree$^\dagger+ \text{Refine}$  & 41.9   & 77.6           & 40.3          & 32.8         & 46.1         &  55.2        & 37.4    & 63.4           & 38.6          & 20.9          & 40.5         & 48.4         & \textbf{24.9}    &   \textbf{46.1}        & \textbf{24.1}           & \textbf{8.6}         & \textbf{28.1}         & \textbf{38.4}        \\\midrule
\multirow{3}{*}{\shortstack{ResNeXt-101- \\FPN \cite{massa2018mrcnn, xie2017aggregated}}}      & No Refine                     & 54.8    & 87.6           & 58.3           & 46.3          & 57.8          & 68.1          & 51.6   & 76.7           & 56.9          & 37.9         & 53.7          & 62.2          & 39.2   & 61.2          & 42.4           & \textbf{20.0}         & 42.3          & 55.7          \\
                          & MOTIF$^\dagger + \text{Refine}$   & \textbf{59.3}   & \textbf{90.6}          & \textbf{64.7}          & \textbf{52.0}         & \textbf{62.2}         &    \textbf{70.6}      & \textbf{54.6}   & \textbf{78.2}           & \textbf{61.1}          & \textbf{41.1}         &    \textbf{56.8}      & \textbf{64.1}         & 39.2   & 61.2           & 42.4           & 19.9         &    42.3      & \textbf{55.8}          \\
                          & VCTree$^\dagger + \text{Refine}$  & 59.0   & 90.4           & 64.2          & 51.7         &  62.0         &  70.4        &  54.3  & 77.9          &  60.4         & 41.0         & 56.4         &    63.8      & 39.2   & 61.2           & 42.4           & 19.9          & 42.3          & 55.7         \\
\bottomrule
\bottomrule
\end{tabular}
}
\caption{\textbf{Segmentation Refinement on MSCOCO. }Standard COCO precision metrics are reported across three tasks and two detector backbones. Task formulation is identical to Table \ref{table:results}. `No Refine' is the baseline where the segmentation masks are obtained from the pre-trained detector. As ground truth masks are unavailable in Visual Genome, evaluation on MSCOCO serves as a proxy.}
\label{table:segmentation}
\vspace{-1.5em}
\end{table*}
\vspace{-0.4em}
\section{Experiments}
\label{sec:experiments}
\vspace{-0.4em}
We perform experiments using two datasets: the Visual Genome Dataset \cite{krishna2017visual} and the COCO dataset \cite{lin2014microsoft}.

\vspace{0.3em}
\noindent\textbf{Visual Genome.} For training and evaluating the scene graph generation performance, we use the Visual Genome dataset \cite{krishna2017visual}. The complete dataset contains $75$k object categories and $37$k relation categories. However, a large fraction of the object annotations have poor quality and nearly $92\%$ of the relation labels have fewer than or equal to $10$ annotations. Therefore, we use the widely adopted prepossessed version of Visual Genome from \cite{xu2017scene}. This subset contains $108$k images across $150$ object categories and $50$ relation labels. We use the original $70$-$30$ split for training and testing, with $5$k images from the training set held out for validation. Images with more that $40$ object bounding boxes are filtered out from the test set due to memory constraints.

\vspace{0.3em}
\noindent
\textbf{MS-COCO} For training and evaluating the segmentation masks, we use the MSCOCO 2017 dataset \cite{lin2014microsoft}, which contains $123$k images, split into $118$k training and $5$k validation images, across $80$ categories. As the ground-truth annotations for the test set are not available, as is common practice, results are reported on the validation set. 
\vspace{-0.3em}
\subsection{Scene Graph Generation Model}
\vspace{-0.3em}
The proposed scene graph segmentation grounding framework is generic and can be easily integrated with various scene graph generation models. In this work we experiment with two scene graph architectures, namely MOTIF \cite{zellers2018neural} and VCTree \cite{tang2019learning}. Both these methods propose architectures that compute global context among objects and relations using stacked recurrent frameworks.

In the case of MOTIF \cite{zellers2018neural}, the object and relation networks (Equation \ref{eq:sg-factor}) are each instantiated by bidirectional LSTMs \cite{hochreiter1997long}. For an image $\mathbf{x}^g_j \in \mathcal{D}^g$, the extracted bounding boxes $\mathbf{B}^g$ are arranged based on their $x$-coordinate position, and passed through the bidirectional LSTM networks. 
%individually when computing object labels, and as pairs when predicting relations. 
Instead of assuming a linear ordering between the objects, VCTree \cite{tang2019learning} generates a dynamic binary tree, with the aim of explicitly encoding the parallel and hierarchical relationships between objects. The object and relation networks are instantiated as bidirectional TreeLSTMs \cite{tai-etal-2015-improved}, which traverse the tree in a top-down as well as bottom-up fashion. 

When augmenting our approach with MOTIF \cite{zellers2018neural} and VCTree \cite{tang2019learning}, we identically replicate the object and relation networks proposed in the respective works. Additional details are provided in the \textbf{appendix}.
% We refer the reader to \cite{zellers2018neural, tang2019learning} for detailed model descriptions.

\subsection{Evaluation}
% We use the following metrics for evaluating the scene graph generation and segmentation performance.
% \vspace{0.05in}
% \\
\noindent
\textbf{Relationship Recall (RR).} To measure the performance of models we use the \textbf{mean Recall @K (mR@K)} metric introduced in \cite{chen2019knowledge, tang2019learning}. The mean Recall metric calculates the recall for predicate label independently across all images and then averages the result. We report the mean Recall instead of the conventional Regular Recall(R@K) due to the long-tail nature of relation labelling in Visual Genome that leads to reporting bias \cite{tang2020unbiased}. Mean Recall reduces the influence of dominant relationships such as \texttt{on} and \texttt{has}, and gives equal weight to all the labels in the dataset.
\vspace{0.05in}
\\
\noindent
\textbf{Zero-Shot Recall (zsR@K).} Introduced and first evaluated on the visual genome dataset in \cite{lu2016visual}, zsR@K computes the Recall@K for subject-predicate-object triplets that are not present in the training data. 

These evaluation metrics are computed for three different sub-tasks: 1) Predicate Classification \textbf{(PredCls)}: predict the relation labels given the ground truth objct bounding boxes and labels; 2) Scene Graph Classification \textbf{(SGCls)}: predict the object and relation labels given the ground truth object bounding boxes; 3) Scene Graph Detection \textbf{(SGDet)}: given an image, predict the entire scene graph.
% \begin{itemize}
%     \item Predicate Classification\textbf{(PredCls)}: Predict the relationship labels given the ground truth object bounding boxes and object labels.
%     \item Scene Graph Classification \textbf{(SGCls)}: Predict the object and realtion labels given the image and ojbect bounding boxes.
%     \item Scene Graph Detection\textbf{(SGDet)}: Given an image, predict the entire scene graph.
% \end{itemize}
\vspace{0.05in}
\\
\noindent
\textbf{Segmentation Precision.} As the Visual Genome dataset \cite{krishna2017visual} does not contain any instance-level segmentation annotations, as a proxy we use the MSCOCO dataset \cite{lin2014microsoft} to measure the performance of the segmentation refinement procedure described in Section \ref{sec:sg-seg-refine}. To make the evaluation similar to scene graph generation, we analogously define three sub-tasks to measure the improvement in the quality of segmentation masks. These sub-tasks, namely \textbf{(PredCls)}, \textbf{(SGCls)}, and \textbf{(SGDet)}, are identical to the ones defined earlier. For each of these sub-tasks, the standard evaluation metrics on COCO are reported \cite{he2017iccv}. 

\vspace{-0.4em}
\subsection{Implementation Details}
\vspace{-0.5em}
\noindent
\textbf{Detector.} For our detector architecture, we use the two-stage Faster-RCNN \cite{ren2015faster} framework. The first stage uses a region proposal network (RPN) to generate class-agnostic object region proposals. The second stage learns to classify the features corresponding to these proposals into objects labels, and simultaneously learns a regressor to refine proposal bounding box. To demonstrate the flexibility of our approach, we experiment with two different backbones within the Faster-RCNN framework: 1) VGG-16 \cite{vgg-16} pre-trained on the ImageNet \cite{imagenet} dataset, and 2) ResNeXt-101-FPN \cite{massa2018mrcnn, xie2017aggregated} backbone pre-trained on the MSCOCO \cite{lin2014microsoft} dataset. We first fine-tune the detector jointly on the Visual Genome and MSCOCO datasets, refining the classifiers and regressors, and simultaneously learning a segmentation network on images in MSCOCO. When training the scene graph generators, the detector parameters are freezed. Note that for the baselines, the detector is fine-tuned only on the Visual Genome, and hence no segmentation is learned.
% We then fine tune the box head on the Visual Genome dataset and freeze the weight for training the baseline scene graph generators. For the proposed segmentation grounded framework we fine tune the mask head and the box head jointly on Visual Genome and COCO.
\vspace{0.05in}
\\
\noindent
\textbf{Scene Graph Models.} For training the scene graph models we use an SGD optimizer with an initial learning rate of $10^{-2}$. Following prior works, we integrate the frequency bias \cite{zellers2018neural} into the training and inference procedure. During inference, in SGDet task, we filter pairs of objects that do not have any bounding box overlap for relation prediction.

\begin{figure*}[t]
    \centering
    \includegraphics[width=.88\linewidth]{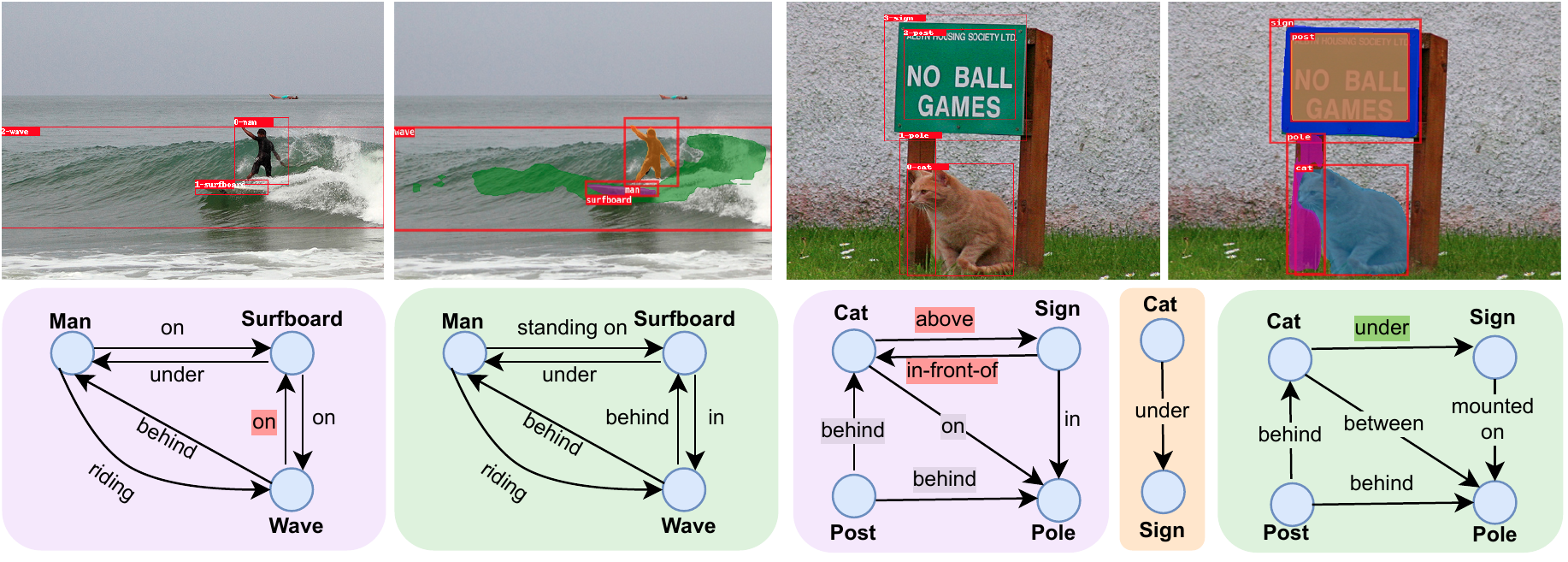}
    \caption{\textbf{Qualitative Results.} Visualizations of scene graphs generated by using VCTree \cite{tang2019learning} (in purple) and our approach augmented to VCTree (in green). The left two images contrast performance on relation retrieval. The right two images contrast performance on zero-shot relation retrieval, with the zero-shot triplet shown in yellow. Our approach additionally generates pixel-level object groundings. }
    % Additional visualizations are shown in the supplementary.}
    \label{fig:qualitative_results}
    \vspace{-1.8em}
\end{figure*}
\vspace{-0.5em}
\section{Results}
\vspace{-0.3em}
\label{sec:results}
\label{sec:ablation}
\noindent
\textbf{Relationship Recall.} We report the mean Recall values comparing the baseline and proposed method in Table \ref{table:results}. To ensure a fair comparison, we additionally report the numbers obtained via re-implementing the baselines. Note that in case of MOTIF \cite{zellers2018neural}, our re-implementation provides significantly higher performance compared to the reported numbers in \cite{zellers2018neural}. For both the MOTIF \cite{zellers2018neural} and VCTree \cite{tang2019learning}, irrespective of the backbone architecture, we observe a consistent improvements in the recall rate across all three tasks when incorporating our proposed approach. 

For MOTIF \cite{zellers2018neural}, we observe an improvement of $7.0\%$ on average at mR@$20$, $50$ and $100$ across all settings and backbones. Specifically, on the VGG backbone \cite{vgg-16}, we obtain a relative improvement of $6.5\%$, $5.3\%$, and $7.7\%$ on mr@$20$ across the three tasks. Similarly, for the ResNeXt-101-FPN \cite{massa2018mrcnn, xie2017aggregated} backbone, we observe relative improvements of $2.8\%$, $11.2\%$, and $10.3\%$ on mr@$20$. Similarly for VCTree \cite{tang2019learning}, an average improvement of $12.6\%$ is observed across tasks and backbones. We attribute the performance improvement to the ability of our model to effectively ground objects and relations to pixel-level regions, thus providing more discriminative features. We provide additional results and individual relation recall comparisons in the \textbf{appendix}.
% $3.9\%$, $11.2\%$ and $10.3\%$ on mR@$20$ for PredCls, SGCls and SGDet tasks. Similarly, for VCTree \cite{tang2019learning}, we observe an average improvement of $?? \%$ across different recall thresholds and tasks. In detail, we obtain relative improvements of $9.9\%$, $15.6\%$ and $ ??\%$ on mR@$20$ for PredCls, SGCls and SGDet tasks. \toadd{VGG comparisons}. We attribute the improvement in performance to more discriminative features obtained the additional zero-shot segmentation grounding which allows for improved representation learning.
\renewcommand{\arraystretch}{1.1}
\begin{table}[t]
\setlength\tabcolsep{2.5pt}
\resizebox{\linewidth}{!}{
\begin{tabular}{@{}cccccc@{}}
\toprule
\toprule
\multirow{2}{*}{Model} & \multirow{2}{*}{Detector} & \multirow{2}{*}{Method}  & PredCls             & SGCls             & SGDet      \\ \cmidrule(l){4-6} 
&                &    & zsr@20/100          & zsr@20/100        & \multicolumn{1}{l}{zsr@20/100} \\ \midrule
\multirow{4}{*}{MOTIF$^\dagger$} & \multirow{2}{*}{VGG-16 \cite{vgg-16}}  & BL     & 1.7/6.7           & 0.2/1.1          & 0.0/0.4                        \\
                        & & SG     & \textbf{3.2/9.3} & \textbf{0.4/1.6} & \textbf{0.1/0.5}               \\ \cdashline{3-6}

                        & \multirow{2}{*}{\shortstack{ResNeXt-101- \\FPN \cite{massa2018mrcnn, xie2017aggregated}}} & BL     & 1.9/7.2           & 0.3/1.2          & 0.0/0.5                        \\
                        & & SG     & \textbf{4.1/10.5} & \textbf{0.8/2.5} & \textbf{0.1/1.0}                          \\ \midrule
\multirow{4}{*}{VCTree$^\dagger$} & \multirow{2}{*}{VGG-16 \cite{vgg-16}} & BL     & 1.8/7.3           & 0.6/1.8          &                                     0.1/0.5                         \\
                        & & SG     & \textbf{3.5/10.2} & \textbf{0.7/2.4} & \textbf{0.3/0.9}                      \\ \cdashline{3-6}
                        & \multirow{2}{*}{\shortstack{ResNeXt-101- \\FPN \cite{massa2018mrcnn, xie2017aggregated}}} & BL     & 1.8/7.1           & 0.4/1.2          & 0.1/0.7                        \\
                        & & SG     & \textbf{4.3/10.6} & \textbf{0.8/2.5} & \textbf{0.3/1.5}                      \\\bottomrule \bottomrule
\end{tabular}
}
\caption{\textbf{Zero-Shot Recall on Visual Genome. } Results are reported for three tasks across two detector backbones. Our approach is augmented to and contrasted against MOTIF \cite{zellers2018neural} and VCTree \cite{tang2019learning}. $\dagger$ denotes our re-implementation of the methods.}
\label{table:zeroshot}
\vspace{-1.8em}
\end{table}
\vspace{0.03in}
\\
\noindent
\textbf{Zero-Shot Recall.} We report the Zero-Shot Recall value zsR@$20$ and zsR@$100$ in Table \ref{table:zeroshot}. We observe an consistent improvement in zero-shot recall when using the proposed scene graph generation framework. Our method outperforms the baselines by an average of $94.5\%$ and $97.9\%$ on MOTIF and VCTree respectively.
\vspace{0.03in}
\\
\noindent
\textbf{Segmentation Accuracy.} As segmentation annotations are not present in Visual Genome \cite{krishna2017visual}, we evaluate our proposed segmentation refinement on the MSCOCO dataset \cite{lin2014microsoft}. This provides a suitable proxy, wherein segmentation improvements on the MSCOCO dataset can be translated, to some degree, on Visual Genome. We report the standard COCO evaluation metrics, namely AP (averaged over IoU thresholds), AP$_{50}$, $AP_{75}$, and AP$_S$, AP$_M$, AP$_L$ (AP at different scales), on three different scene graph evaluation tasks in Table \ref{table:segmentation}. `No Refine' acts as a strong baseline, wherein the segmentation masks are generated from the pretrained detector. It is evident that our proposed segmentation refinement improves on mask quality across tasks and detector backbones. As the ground truth bounding boxes and labels are available for the Predicate Classification task, the observed improvement here is the largest ($34.6\%$ higher AP on VGG). Analogously, the observed improvement on the Scene Graph Generation (SGDet) task is the lowest ($7.3\%$ higher AP on VGG) as any errors made by the pretrained detector are forwarded to the scene graph network. Additionally, we do not observe any noticeable improvement for the SGDet task when ResNeXt-101-FPN \cite{massa2018mrcnn, xie2017aggregated} is used as the detector backbone. We believe this is a direct consequence of the backbone using feature pyramid networks (FPNs) \cite{lin2017feature} to compute feature representations for a given image. As FPNs effectively capture global context using lateral connections, the object detector provides much richer object representations. This makes the context aggregation in the scene graph network redundant, making refining segmentation masks harder.
\renewcommand{\arraystretch}{1.1}
\begin{table}[t]
\setlength\tabcolsep{2.5pt}
\resizebox{\linewidth}{!}{
\begin{tabular}{@{}lcccc@{}}
\toprule
\toprule
                     & \multicolumn{4}{c}{Scene Graph Classification} \\ \cmidrule(l){2-5} 
Ablation             & mR@20         & mR@50         & mR@100    & zsr@20/100      \\ \midrule
Base                 & 8.1          & 9.9          & 10.6     & 0.3/1.2    \\
Joint       & 8.5           & 10.5          & 11.1      & 0.4/1.5    \\
% Joint$_{schedule}$ & 9.3 & 11.4 & 12.2 & 0.8/2.3 \\
Joint + OG  & 9.0           & 11.1          & 11.8      & 0.6/2.1    \\
Joint + OG + EG$_{avg}$ &9.1           &11.4               &12.2       & 0.8/2.4         \\
Joint + OG + EG$_{union}$        & 9.1          & 11.3         & 12.2  & 0.7/2.4 \\
Joint + OG + EG$_{Gaussian}$   & 9.3           & 11.5         & 12.2   & 0.7/2.3      \\
Final Model & \textbf{9.4}          & \textbf{11.6}         & \textbf{12.3}     & \textbf{0.8/2.5} \\ \bottomrule
\bottomrule
\end{tabular}
}
\caption{\textbf{Ablation. } Mean Recall (mr) and Zero-shot Recall (zsr) are reported. VCTree \cite{tang2019learning} is the base architecture for all methods. Please refer Section \ref{sec:ablation} for model definitions.}
\label{table:ablation}
\vspace{-1.8em}
\end{table}
\vspace{0.03in}
\\
\noindent
\noindent\textbf{Ablation. }We conduct an ablation study over the various components in our model using VCTree as the base architecture. All models are trained with the ResNeXt-101-FPN \cite{massa2018mrcnn, xie2017aggregated} backbone. The results on the Scene Graph Classification task are shown in Table \ref{table:ablation}. `Base' is defined as the vanilla VCTree model learned over the detector trained only on the Visual Genome dataset. That is, `Base' does not utilize our proposed joint training approach. To understand the effect our joint detector pre-training has on the overall performance, we define `Joint' as the VCTree model learned over this jointly pre-trained detector. It can be seen that just the joint pre-training of the detector provides considerable improvements ($5\%$ on mR@20). 

We incrementally add components of our proposed approach to the `Joint' detector to better highlight their importance. `Joint + OG' is defined as the model that uses the jointly trained detector and the object grounding mechanism described in Section \ref{sec:sg-object}. Similarly, `Joint + OG + EG$_{x}$' describes the model that additionally uses our proposed relation grounding mechanism defined in Section \ref{sec:sg-relation}. The subscript $x$ in $\text{EG}_x$ refers to the type of attention mechanism used to combine the segmentation masks for a pair of objects. We experiment with averaging ($avg$), taking the logical or ($union$), and the proposed Gaussian attention ($Gaussian$). Finally, our complete model with the additional segmentation mask refinement (Section \ref{sec:sg-seg-refine}) is denoted as `Final Model'. From Table \ref{table:ablation} it can be seen that using both object and relation grounding helps with performance, and using a Gaussian attention mechanism is superior to other alternatives. Additionally, fine-tuning the segmentation masks not only helps improve its quality, but also provides better scene graph generation performance.
\vspace{0.03in}
\\
\noindent
\textbf{Qualitative Results. } We qualitatively contrast the performance of the VCTree model \cite{tang2019learning} augmented with our proposed approach against its vanilla counterpart in Figure \ref{fig:qualitative_results}. The two images on the left show results from the relation retrieval task. Our approach (in green) predicts more granular and spatially informative relationships \texttt{standing on} and \texttt{behind}, as opposed to the the baseline (in purple) which is heavily biased towards the more common and less informative relation \texttt{on}. The two images on the right highlights the ability of our approach to generalize in zero-shot scenarios. As the triplets of \texttt{cat} with \texttt{sign} are absent from the training dataset, the baseline approach (in purple) defaults to predicting incorrect relations of \texttt{above} and \texttt{in-front-of}. On the contrary, our approach accurately predicts the correct relation \texttt{under}.
\vspace{-0.5em}
\section{Related Work}
\vspace{-0.5em}
\noindent
\textbf{Scene Graph Generation.}
Scene graph generation is a popular topic % has recently attracted more attention 
in the vision community \cite{knyazev2020graph, li2018fnet, Lin_2020_CVPR, newell2017pixels, tang2020unbiased, xu2017scene, yang2018graph, ZareianKC20, ZareianWYC20, zellers2018neural}. Scene graph generation methods can be roughly categorized into two; one-stage and two-stage methods. First introduced in \cite{xu2017scene}, they proposed an iterative message passing module to refine the features of node and edge prior to classification. Subsequent works proposed use of different architectures such as Bi-directional LSTM \cite{zellers2018neural}, Tree-LSTM \cite{tang2019learning}, Graph Neural Networks \cite{yang2018graph} and novel message passing algorithm \cite{li2018factorizable,qi2019attentive} for representation learning. While improved context aggregation can lead to better scene graph performance, recent works have focused more on alleviating issue rising from long tail distribution of relation labels. Tang \etal \cite{tang2020unbiased} propose the use of a causal inference framework to debias the prediction of model obtained from biased training. Knyazev \etal \cite{knyazev2020graph} in their work propose a graph density aware loss akin to focal loss to address the imbalance in the Visual Genome dataset.  

\vspace{0.1em}
\noindent
\textbf{Zero-Shot Segmentation.} Zero-shot learning is an active area of research in computer vision \cite{geng2020recent, xian2017zero}. The sub-field of zero-shot segmentation, however, is relatively recent \cite{bucher2019zero,hu2020uncertainty, Kato_2019_ICCV,khandelwal2021unit,zhao2017open}. A majority of work in this area has been on zero-shot semantic segmentation \cite{bucher2019zero, hu2020uncertainty,Kato_2019_ICCV,xian2019semantic,zhao2017open}, where the aim is correct categorize each pixel in an image. Zhao \etal \cite{zhao2017open} propose the open-vocabulary scene parsing task, wherein hypernym/hyponym relations from WordNet \cite{miller1995wordnet} are leveraged to build label relationships, and subsequently segment classes. 
%Xian \etal \cite{xian2019semantic} utilize class-level semantic information to segment classes in both zero and few-shot settings.
Bucher \etal \cite{bucher2019zero} combine visual-semantic embeddings along with a generative model and classifier to obtain masks for unseen classes. Kato \etal \cite{Kato_2019_ICCV} leverage a semantic to visual variational mapping over class labels, along with a data driven distance metric, to generate zero-shot segmentation masks. Hu \etal \cite{hu2020uncertainty} introduce uncertainty aware losses to mitigate the effect of noisy training examples for robust semantic segmentation. For the task of zero-shot instance-level segmentation, Khandelwal \etal \cite{khandelwal2021unit} leverage linguistic and visual similarities to learn a transformation over segmentation heads from classes with abundant annotations to classes with zero/few annotations.

\vspace{0.1em}
\noindent
\textbf{Multi-task Learning.} Multi-task Learning (MTL) involves optimizing for several tasks simultaneously and transferring information across task for improved performance \cite{Multitask}. Most methods in MTL \cite{chu2015multi,long2015learning,zhang2014facial} learn a shared representation of layers along with multiple independent classifiers. Another line of work involves explicitly modelling the the relation between tasks, either by grouping \cite{jacob2008clustered, kang2011learning, kumar2012learning} or in the form of task-covariance \cite{ciliberto2015convex,evgeniou2004regularized,zhang2010learning}. For a comprehensive survey, we refer the reader to \cite{zhang2017survey}.

\vspace{-0.6em}
\section{Conclusion}
\vspace{-0.6em}
% We present a novel framework to ground objects and relations present in a scene graph to pixel-level regions 
We present a novel model-agnostic framework for segmentation grounded scene graph generation. Contrary to tradition scene graph generation frameworks that grounds object in a scene graph to bounding boxes, our proposed methodology allows for a more granular pixel-level grounding, obtained via a zero-shot transfer mechanism. Our proposed framework leverages these groundings to provide significant improvements across various scene graph prediction tasks, irrespective of the architecture it is augmented to. Finally, we highlight the benefits of simultaneously optimizing the tasks of scene graph and segmentation generation, which leads to improved performance on both.
% The benefits of our proposed framework are demonstrated by augmenting it to existing architectures,  significant improvements are observed across various scene graph  and detector backbones. Finally, we also show how the representations learned for scene graph generation can help improve segmentation performance on MSCOCO dataset.
\begin{ack}
This work was funded, in part, by the Vector Institute for AI, Canada CIFAR AI Chair, NSERC CRC and an NSERC DG and Accelerator Grants. This material is based upon work supported by the US Air Force Research Laboratory (AFRL) under the DARPA Learning with Less Labels (LwLL) program (Contract No.FA8750-19-C-0515) \footnote{The views and conclusions contained herein are those of the authors and should not be interpreted as representing the official policies or endorsements, either expressed or implied, of DARPA or the U.S. Government.}.

Hardware resources used in preparing this research were provided, in part, by the Province of Ontario, the Government of Canada through CIFAR, and companies sponsoring the Vector Institute\footnote{\url{ www.vectorinstitute.ai/\#partners}}. 
Additional support was provided by JELF CFI grant and Compute Canada under the RAC award. 

Finally, we sincerely thank Bicheng Xu for valuable feedback on the paper draft. 
\end{ack}

{\small
\bibliographystyle{ieee_fullname}
\bibliography{egbib}
}

\end{document}